\documentclass[letterpaper, 10 pt, conference]{ieeeconf}  

\IEEEoverridecommandlockouts              
\usepackage{amsmath} 
\usepackage{amssymb}  
\usepackage{mathtools}
\usepackage{multirow}
\usepackage{cite}
\usepackage{graphicx}
\DeclarePairedDelimiter{\norm}{\lVert}{\rVert}
\usepackage{xcolor}
\usepackage{hyperref}

\title{\LARGE \bf
An Open Tele-Impedance Framework to Generate Data for Contact-Rich Tasks in Robotic Manipulation
}

\author{Alberto Giammarino~\IEEEmembership{Member, IEEE}, Juan M. Gandarias,~\IEEEmembership{Member, IEEE} and Arash Ajoudani,~\IEEEmembership{Member, IEEE}
\thanks{The authors are with the HRI$^{2}$ Lab, Istituto Italiano di Tecnologia, Genoa, Italy.
        {\tt\small \{alberto.giammarino, juan.gandarias, arash.ajoudani\}@iit.it}
        }
\thanks{This work was supported by the ERC-StG Ergo-Lean (Grant Agreement No.850932).}
        }

\begin{document}

\maketitle
\thispagestyle{empty}
\pagestyle{empty}

\begin{abstract}

Using large datasets in machine learning has led to outstanding results, in some cases outperforming humans in tasks that were believed impossible for machines. However, achieving human-level performance when dealing with physically interactive tasks, e.g., in contact-rich robotic manipulation, is still a big challenge. It is well known that regulating the Cartesian impedance for such operations is of utmost importance for their successful execution. Approaches like reinforcement Learning (RL) can be a promising paradigm for solving such problems. More precisely, approaches that use task-agnostic expert demonstrations to bootstrap learning when solving new tasks have a huge potential since they can exploit large datasets. However, existing data collection systems are expensive, complex, or do not allow for impedance regulation. This work represents a first step towards a data collection framework suitable for collecting large datasets of impedance-based expert demonstrations compatible with the RL problem formulation, where a novel action space is used. The framework is designed according to requirements acquired after an extensive analysis of available data collection frameworks for robotics manipulation. The result is a low-cost and open-access tele-impedance framework\footnote{\url{https://gitlab.iit.it/hrii-public/teleimpedance-m5core2}} which makes human experts capable of demonstrating contact-rich tasks.

\end{abstract}

\section{Introduction}
\label{sec:introduction}

Performing contact-rich manipulation tasks is still a big challenge in robotics. Solving them would unlock the use of robots for applications like logistics and assembly, where their use can have a strong impact. Different aspects make contact-rich manipulation particularly hard to solve. The dynamics of the system around contacts are difficult to model, and small modeling errors can induce task failure. Moreover, perception of the scene is hard due to the frequent co-existence of multiple different objects along with the manipulating hand in a small region.

Imitation Learning (IL)~\cite{hussein2017imitation} is a viable approach for such problems, and one widely used IL approach is behavioral cloning (BC)~\cite{torabi2018behavioral}. Although it has been successfully applied, it suffers from distributional shift~\cite{ross2011reduction}, meaning that it struggles outside the manifold of the demonstration data. 
An alternative IL approach is Inverse Reinforcement Learning (IRL)~\cite{ng2000algorithms}. Despite the appealing idea, the framework makes the limiting assumption of having expert task-specific demonstrations and the joint optimization of policy and reward renders the training process difficult to stabilize.

\begin{figure}
    \centering
    \includegraphics[width=0.65\columnwidth]{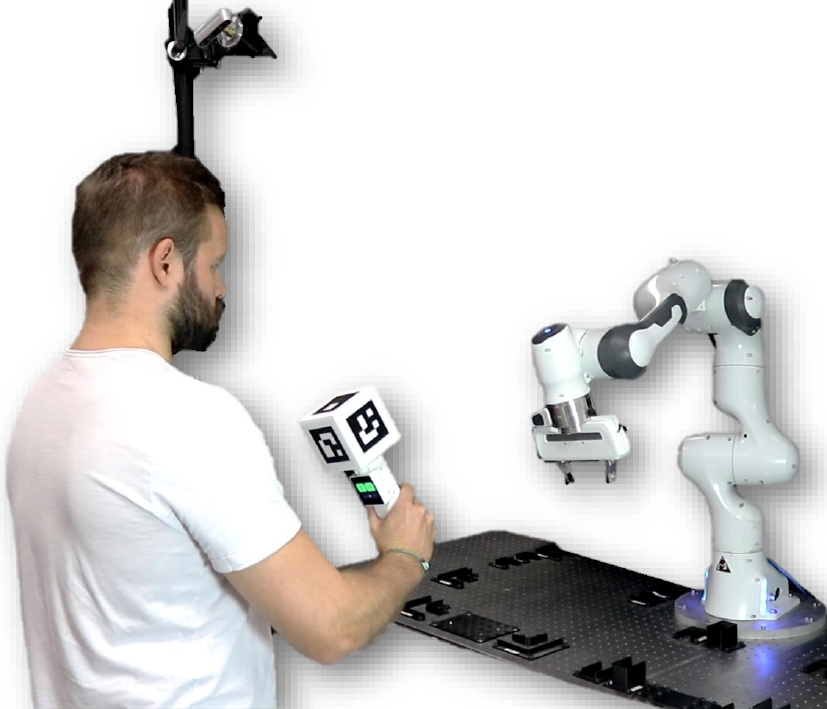}
    \caption{A human expert teleoperating a robotic manipulator thanks to the proposed open tele-impedance framework. A video of the experimental session and a demonstration of the framework using a real Franka Emika Panda robotic manipulator is included\protect\footnotemark}
    \label{fig:first_figure}
    \vspace{-0.7cm}
\end{figure}
\footnotetext{\url{https://youtu.be/60X74kCUTes}}

Another attractive framework is Reinforcement Learning (RL), where the agent learns to accomplish a task by trial and error, maximizing an expected future cumulative reward. 
In the context of robotic manipulation, a natural reward choice is sparse and binary (i.e., +1 when the task succeeds, 0 otherwise). Before the agent is able to receive useful feedback, it has to explore the state-action space until task completion, which makes RL algorithms extremely sample inefficient for this application. One possible solution to this problem is reward shaping. However, designing a reward is often counterintuitive, requires handcrafting, might lead to unexpected undesired behaviors, and is prone to local optima~\cite{ng1999policy}.

An interesting alternative to reward shaping is the integration of task demonstrations into RL algorithms (Fig. \ref{fig:first_figure}). The additional knowledge encoded into the demonstrations is used to bootstrap learning, and speed it up. 
A class of these approaches leverages task-agnostic demonstrations to learn a new task. In this case, demonstrations are used as generic prior experience to learn aspects of environment and task. We deem that these methods are among the most promising ones since they can make use of large datasets, which have resulted in remarkable results in other machine learning fields. Moreover, these methods resemble how humans learn new tasks. Prior experience allows them to narrow down the search space and as a consequence, they are able to master a new task after a few trials. In this paper, we refer to these approaches as task-agnostic demonstration-based RL.

It is well-known that humans regulate the impedance of their arms according to the requirement while performing contact-rich tasks~\cite{ajoudani2012tele}. In~\cite{martin2019variable}, it is shown that formulating the action space as Cartesian impedance and end-effector pose improves the sample efficiency of model-free RL algorithms with respect to common formulations. Although that, none of the task-agnostic demonstration-based RL approaches discussed in Section \ref{subsec:RL_approaches} uses this action space formulation. This might be due to the lack of a suitable data collection system.

In this paper, we design a system to ease and enable the collection of motion and impedance expert trajectories for contact-rich manipulation with Cartesian impedance-controlled robots. More specifically, we target tasks that require impedance adaptation to be successfully fulfilled. We gather insights from related works on data collection systems and task-agnostic demonstration-based RL approaches to understand which the requirements for such a system should be. Finally, we design and test the devised framework.


\section{Related Work}
\label{sec:rel_work}

\subsection{Task-agnostic Demonstration-based RL}
\label{subsec:RL_approaches}

The approaches discussed here integrate a dataset of task-agnostic interactions (i.e., state-action sequences) into RL to learn a new contact-rich task. In~\cite{singh2020parrot} the authors use generative modeling and representation learning to learn a model that maps a random (latent) variable to meaningful action distributions conditioned on the current state. Demonstrations come from a mix of different successful tasks, where meaningful behaviors are executed. Learning a policy in the space of the latent variable becomes easier because random actions are more likely to correspond to user behaviors.
A similar approach is adopted in~\cite{pertsch2020accelerating}, where exploration is guided using a learned prior distribution over skills (i.e., action sequences) conditioned on the current state. The same authors in~\cite{pertsch2021guided} improve their previous framework by adding a few task-specific demonstrations in the dataset. In~\cite{gupta2019relay} demonstrations of semantically meaningful behaviors coupled with data-relabeling are used to initialize goal-conditioned hierarchical policies. During RL for learning a new task, the loss is augmented to keep the policies close to the behavior encoded in the dataset. 

Some works integrate demonstrations in the offline (or batch) RL framework~\cite{levine2020offline}. The authors of~\cite{singh2020cog} enhance the generalization capability of an offline RL algorithm, i.e., conservative Q-Learning (CQL)~\cite{kumar2020conservative}, using a dataset of prior experience unrelated to the new task. In~\cite{siegel2020keep} a behavior prior is learned from an imperfect dataset of multiple tasks and is used to avoid that the policy chooses actions not supported by the data.

\subsection{Data Collection Systems}
\label{subsec:data_collect_sys}

Machine-generated demonstrations are a tempting idea to generate datasets. In~\cite{siegel2020keep} an agent is trained with MPO~\cite{abdolmaleki2018maximum} and roll-outs of the final policy are collected to build a dataset. Another way to collect machine-generated demonstrations is through the use of scripted policies, as it is done in~\cite{singh2020parrot} and~\cite{singh2020cog}. However, humans are more suitable for collecting a wide set of semantically meaningful and natural behaviors, which is harder to guarantee for machine-generated data.

The works in~\cite{pertsch2021guided} and ~\cite{gupta2019relay} use MuJoCo HAPTIX~\cite{kumar2015mujoco} as data collection system. The authors of~\cite{kumar2015mujoco} propose to augment the MuJoCo simulator~\cite{todorov2012mujoco} with a system that enables teleoperation using real-time motion capture (MoCap) of arm and hand movements, coupled with a stereoscopic monitor for visual feedback. This is combined with head-tracking to show the virtual environment always from the human viewpoint. This system is employed also in~\cite{pertsch2020accelerating} and~\cite{nair2020accelerating}, where the former uses a dataset created in~\cite{fu2020d4rl} while the latter exploits behavioral cloning to augment the dataset collected in~\cite{rajeswaran2017learning} with an updated version of MuJoCo HAPTIX.

\begin{figure*}
    \centering
    \includegraphics[width=0.25\textwidth]{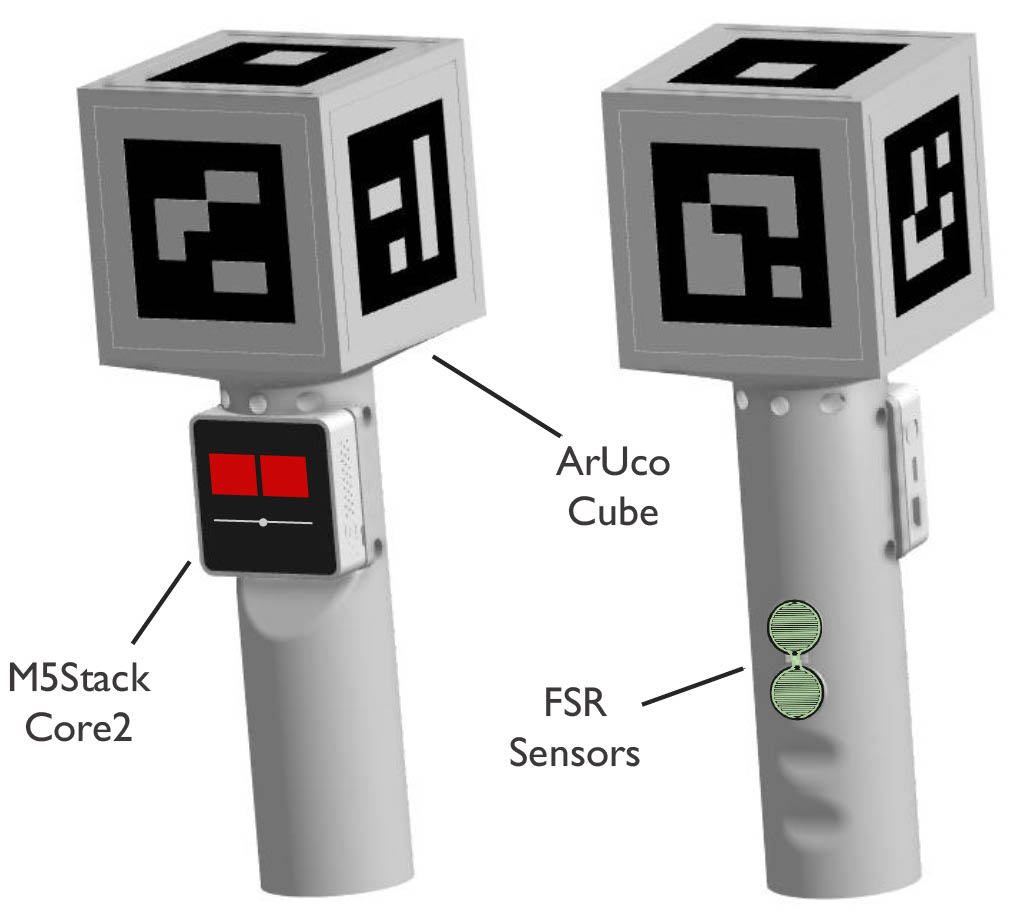}
    \includegraphics[width=0.20\textwidth]{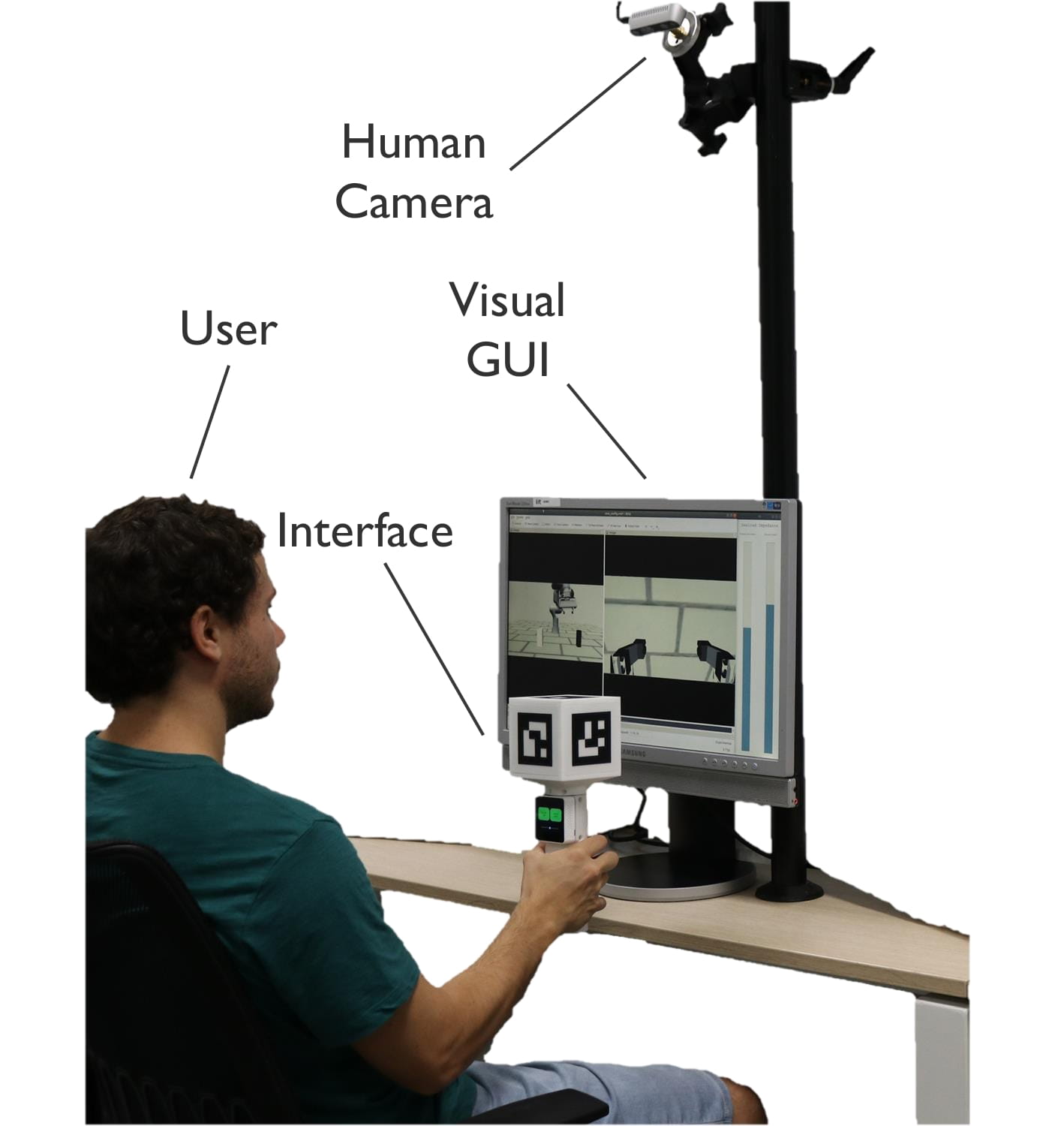}
    \includegraphics[width=0.37\textwidth]{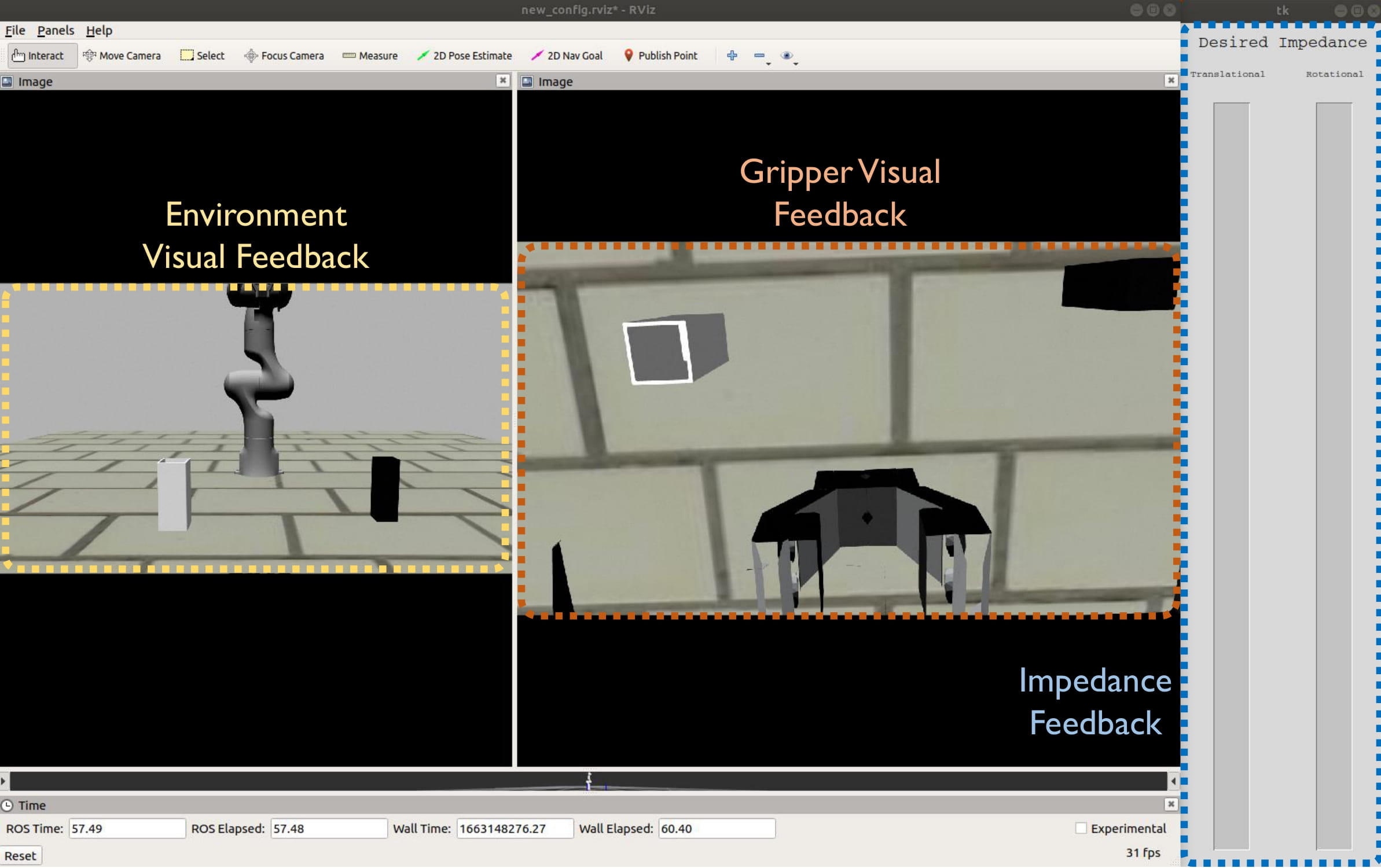}\\
    \hspace{-1.3cm} (a) \hspace{4cm}  (b) \hspace{5cm} (c)
    \caption{Design concept of the proposed Tele-impedance System. (a) Tele-impedance interface, (b) Human side, (c) Human visual feedback composed of (from left to right) static camera view showing robot side and its workspace, wrist-mounted camera view, and feedback of the commanded translational and rotational impedance.}
    \label{fig:data_collect_sys_design}
    \vspace{-0.7cm}
\end{figure*}

The authors of~\cite{zhang2018deep} aim at building an inexpensive teleoperation framework that allows intuitive robotic manipulation and collection of high-quality demonstrations. They come up with a system that uses a VR headset to let the user perceive the environment through the robot's sensor space and control it with motion-tracked VR controllers. A similar system is used also in~\cite{nair2018overcoming}, where the demonstrations are collected in a virtual environment, and the demonstrator sees a rendering of the same observations of the agent and records actions through an HRC Vive interface. In contrast, in order to use kinesthetic teaching while avoiding human presence in the robot scene, in~\cite{vecerik2017leveraging} a robot kinesthetically force controlled by a human is used to teleoperate the same robot type in simulation or in the real world.

In~\cite{mandlekar2018roboturk} RoboTurk is developed, a crowdsourcing platform for 6 degrees-of-freedom (DoFs) trajectory-based teleoperation. It allows the users to provide task demonstrations in a virtual environment through their own mobile devices. The system has the advantages of using a ubiquitous interface, i.e., a mobile device, while being intuitive and sufficiently precise.

\section{Methodology}
\label{sec:methods}

\subsection{Data Collection System Requirements}
\label{subsec:requirements}

The problem tackled in this work can be stated as that of creating a data collection system for obtaining expert demonstrations $\tau_E$ of state-action trajectories such that i) it is as suitable as possible for task-agnostic demonstration-based RL methods, ii) the VICES action space~\cite{martin2019variable} is chosen in the problem formulation, and iii) the system is usable by the widest possible range of researchers. 

To fulfill the three points aforementioned, we formulate the following requirements developed according to insights gathered from the works in Section \ref{sec:rel_work}:

\begin{itemize}
    \item \textit{Simplicity}. The larger and more various are the datasets, the more the RL approaches considered will benefit from them. Thus, the system should be easy and fast to use.
    \item \textit{Accessibility}. The system should be as much accessible as possible, meaning that the tools used for implementing it should be ubiquitous, low-cost, and thus easily accessible.
    \item \textit{Compatibility}. RL is often cast as the problem of learning visuomotor policies, i.e. motor policies conditioned on images. In order to make the expert demonstrations usable, they should be compatible with the interactions of the agent with the environment during RL training. Thus, firstly visual artifacts should be avoided in the context of visuomotor policy learning. That is, the humans should not be present in the scene during the demonstration (e.g., in kinesthetic teaching) since they will not be during agent training. Secondly, human visual feedback and the images provided as input to the robot's policy should coincide in order to avoid that the decisions taken by the human cannot be correlated to the input image gotten by the robot (e.g., when the robot's view is obstructed but the human one is not).
    \item \textit{Intuitiveness}. Human demonstrations have the advantage of representing more natural and meaningful behaviors than machine-generated demonstrations, characteristics that can potentially drive the success of the RL approaches considered. However, a lack in intuitiveness can easily cancel out this advantage. Both the way demonstrations are provided and the feedback gotten by the demonstrator play a key role in ensuring intuitiveness.
    \item \textit{Relaxed Precision}. In the RL approaches considered, expert demonstrations can provide a coarse indication of what actions are worth to be taken in a state since they aid learning for a new task. Therefore, framework precision can be slightly relaxed as long as successful demonstrations can be still provided.
    \item \textit{VICES Action Space}. The framework should be able to command desired poses and Cartesian impedances to the robot.
\end{itemize}

\subsection{Data Collection Framework Design}
\label{subsec:data_collection_framework_design}

Our data collection framework is depicted in Fig. \ref{fig:data_collect_sys_design}. In particular, Fig. \ref{fig:data_collect_sys_design}(b) shows all the components of the framework, Fig. \ref{fig:data_collect_sys_design}(a) enlarges and highlights the components of the physical interface used by the human to teleoperate the robot and Fig. \ref{fig:data_collect_sys_design}(c) depicts the the visual feedback provided to the human.

\subsubsection{Desired Pose command}

The desired end-effector pose can be commanded to the robot through a MoCap system that tracks directly the human hand, gaining intuitiveness. Technologies based on wearable sensors, e.g., XSens based on Inertial Measurement Units (IMUs), while highly precise usually are not easily accessible (i.e. expensive). Contrarily, technologies based on cameras exist which are open-source, have acceptable precision, and require only cheap cameras to work. In addition, accurately placing sensors and calibrating them takes additional time, at the expense of simplicity. Therefore, we opt for tracking a polyhedron of ArUco markers similarly to the work in~\cite{wu2017dodecapen}, using an RGB camera and the open-source ArUco detect algorithm\footnote{\url{http://wiki.ros.org/aruco_detect}}. The polyhedron of ArUco markers is attached on top of an interface that is tightly grasped by the human during teleoperation, both are 3D printed.

Let $\mathcal{C}$, $\mathcal{A}_i$, $\mathcal{I}$ and $\mathcal{W}_h$ be respectively the camera frame, the $i$-th ArUco marker frame ($i=1,\dots ,n$ with $n+1$ being the number of faces of the polyhedron), the interface frame located where the human grasps it, and the human world frame. The polyhedron of ArUco markers is rigidly attached on top of the interface far enough from the human grasping point. This allows for a) making the ArUco markers more visible to the camera and b) knowing the (constant) homogeneous transformation $\boldsymbol{T}_{\mathcal{A}_i\mathcal{I}}$ from each ArUco frame to the interface frame. Considering the $i$-th ArUco marker, the ArUco detect algorithm provides an estimation of $\boldsymbol{T}_{\mathcal{C}\mathcal{A}_i}$, while the constant transformation $\boldsymbol{T}_{\mathcal{W}_h\mathcal{C}}$ is known once the camera has been fixed and calibrated. Therefore, the $i$-th ArUco marker provides the homogeneous transformation $\boldsymbol{T}^i_{\mathcal{W}_h\mathcal{I}}$ obtained as:
\begin{equation}
    \small
    \boldsymbol{T}^i_{\mathcal{W}_h\mathcal{I}} = \boldsymbol{T}_{\mathcal{W}_h\mathcal{C}}\boldsymbol{T}_{\mathcal{C}\mathcal{A}_i}\boldsymbol{T}_{\mathcal{A}_i\mathcal{I}}.
\end{equation}
Since only one camera is used, one ArUco marker is not sufficient to command the desired pose when the desired orientation changes. The polyhedron of ArUco markers can circumvent this problem and can make the pose tracking more robust. 

Let $_{\mathcal{C}}\boldsymbol{p}_{\mathcal{C}\mathcal{A}_i}$ be the position vector that goes from the camera to the $i$-th ArUco marker expressed in the camera frame, and let $\boldsymbol{R}_{\mathcal{C}\mathcal{A}_i}$ be the rotation matrix describing the orientation of the $i$-th ArUco marker with respect to the camera. The third column of $\boldsymbol{R}_{\mathcal{C}\mathcal{A}_i}$ is $_{\mathcal{C}}\boldsymbol{e}^{\mathcal{A}_i}_z$, that is the $z$-axis of frame $\mathcal{A}_i$ expressed in frame $\mathcal{C}$, which is the one coming out perpendicularly from the ArUco surface. In agreement with the considerations in~\cite{schweighofer2006robust} regarding the pose ambiguity problem and accordingly to the heuristic observations, we consider $\boldsymbol{T}_{\mathcal{C}\mathcal{A}_i}$ more reliable when the angle between $_{\mathcal{C}}\boldsymbol{p}_{\mathcal{C}\mathcal{A}_i}$ and $-_{\mathcal{C}}\boldsymbol{e}^{\mathcal{A}_i}_z$ is small. We compute the cosine similarity between $_{\mathcal{C}}\boldsymbol{p}_{\mathcal{C}\mathcal{A}_i}$ and $-_{\mathcal{C}}\boldsymbol{e}^{\mathcal{A}_i}_z$ as:
\begin{equation}
    \small
    \alpha_i = \frac{-_{\mathcal{C}}\boldsymbol{e}^{\mathcal{A}_i}_z \cdot _{\mathcal{C}}\boldsymbol{p}_{\mathcal{C}\mathcal{A}_i}}{\norm{-_{\mathcal{C}}\boldsymbol{e}^{\mathcal{A}_i}_z}\norm{_{\mathcal{C}}\boldsymbol{p}_{\mathcal{C}\mathcal{A}_i}}} = \cos{\theta_i},
\end{equation}
where $\theta_i$ is the angle between the two vectors and $\alpha_i \in [0,1]$. Then, $\boldsymbol{T}_{\mathcal{W}_h\mathcal{I}}$ is computed as:
\begin{equation}
    \small
    \boldsymbol{T}_{\mathcal{W}_h\mathcal{I}} = \sum_{i | \alpha_i > \alpha_{thr}} \boldsymbol{T}^i_{\mathcal{W}_h\mathcal{I}} \alpha_i,
\end{equation}
where $\alpha_{thr}$ is a threshold below which $\boldsymbol{T}_{\mathcal{C}\mathcal{A}_i}$ is not considered. The higher $n$, the more robust the pose tracking of the interface is. In this work we use $n=5$, i.e. the polyhedron used is a cube. As we will demonstrate in Section \ref{sec:experiments_results} this choice is enough in our case to have a sufficiently precise pose estimation. We choose $\theta_{thr} = \arccos (\alpha_{thr}) = 50^{\circ}$ resulting in $\alpha_{thr} \approx 0.64$.

\subsubsection{Desired Impedance command}

Previous work has tried to estimate the human impedance to transfer it to the robot, in order to make commanding an impedance as natural, intuitive, and precise as possible~\cite{ajoudani2012teleimpedance_conf, wu2020intuitive}. However, for estimating it precisely, sensors that measure human muscular activity, e.g., electromyographic sensors (EMGs), are needed. These sensors are not easily accessible, require calibration, and should be worn by humans. Instead, we decide to use resistive buttons. The pressure applied to them deforms an internal resistance and causes a voltage signal change, which in turn can be mapped to a change in the desired impedance. 

Formally, we command a desired diagonal stiffness $\boldsymbol{K}_d = diag \{\boldsymbol{k}_d\} \in \mathbb{R}^{6\times 6}$, bounded between a minimum and a maximum value, and the corresponding desired damping $\boldsymbol{D}_d = diag \{\boldsymbol{d}_d\} \in \mathbb{R}^{6\times 6}$ is obtained through double diagonalization formula:
\begin{equation}
    \small
    \boldsymbol{d}_d = 2 \cdot 0.707 \cdot \sqrt{\boldsymbol{k}_d}.
\end{equation}
The buttons are connected to the interface that sends the desired Cartesian impedance  to the robot controller.

In the process of deciding how many impedance dimensions the human expert should be able to command independently, which corresponds to the number of buttons to include on the physical interface, we observe that
too many buttons would degrade the intuitiveness of the framework. Hence, we decide to consider a trade off between intuitiveness and complexity and included only two buttons, one for translational and one for rotational impedance.

\subsubsection{Feedback Provided to Human and Other Commands}
When learning visuomotor policies for manipulation tasks, a viable choice is to use a wrist-mounted camera and a static camera recording the whole task workspace~\cite{mandlekar2021matters}.
The views of these cameras are provided as feedback to the human. With this choice, the system might lose intuitiveness (the cameras could provide a non-functional view of the scene for the task purpose) while gaining compatibility (human demonstrated actions are conditioned on the robot state). Another important feedback is the interaction force of the robot end-effector with the environment and the commanded impedance. The former is provided by means of vibrotactile feedback generated by the interface while the latter through visual feedback of progress bars on the screen.

\begin{figure}
    \centering
    \includegraphics[width = 0.8\columnwidth]{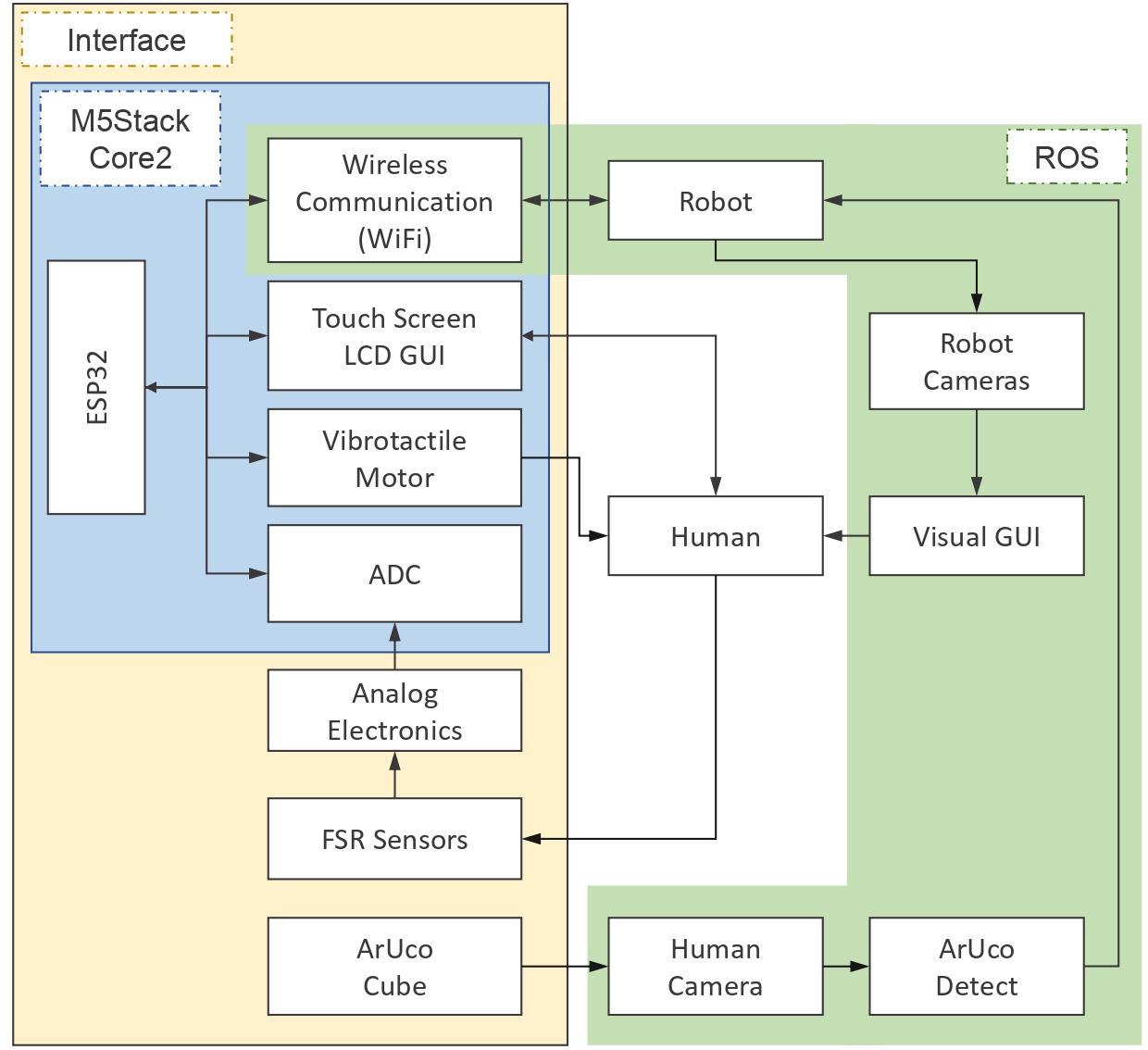}
    \caption{Block diagram of the tele-impedance framework. The human-robot communication is obtained thanks to the tele-impedance interface (see Fig.~\ref{fig:data_collect_sys_design}) and visual GUI. The communication between the different hardware elements according to the connections in the figure is done in ROS.}
    \label{fig:diagram}
    \vspace{-0.7cm}
\end{figure}

Additionally, the interface provides a touch screen where we implement:
\begin{itemize}
    \item A button to command the gripper (open/close).
    \item A button to deactivate and activate the teleoperation.
    \item A slider to command the scaling factor between the leader and the follower movements.
\end{itemize}

\subsection{Tele-Impedance Framework Components}

According to the aforementioned design description, the tele-impedance framework shown in Fig.~\ref{fig:data_collect_sys_design} is implemented. The diagram of Fig.~\ref{fig:diagram} illustrates how all the final components of the framework interact with each other. The interface uses an M5Stack Core2\footnote{\url{https://docs.m5stack.com/en/core/core2}} as a human-robot communication unit. An ESP32 microprocessor manages the hardware components of the interface and communicates with the robot via WiFi. The human can regulate the impedance of the robot with two Force Sensor Resistor (FSR). A module formed by two voltage dividers with resistors of 3.3k$\Omega$ is used as the analog electronics of the FSR. The human can also communicate with the robot through the M5Stack Core2 touch screen. In addition, the human receives feedback of the task through the visual GUI, and vibrotactile feedback of the force measure at the end-effector of the robot. More details about the hardware components and assembly instructions are included in the open repository of the framework.

\subsection{Data Logging and Simulation Environment}
\label{subsec:data_logging}
An important feature for data collection systems for learning applications is the autonomous data logging. We include such feature in the code we open source, so that states-action trajectories can be readily available once the demonstration is terminated. In addition, we provide the simulation environment used for the experiment presented in the next section.

\begin{figure*}
    \centering
    \includegraphics[width=0.89\textwidth]{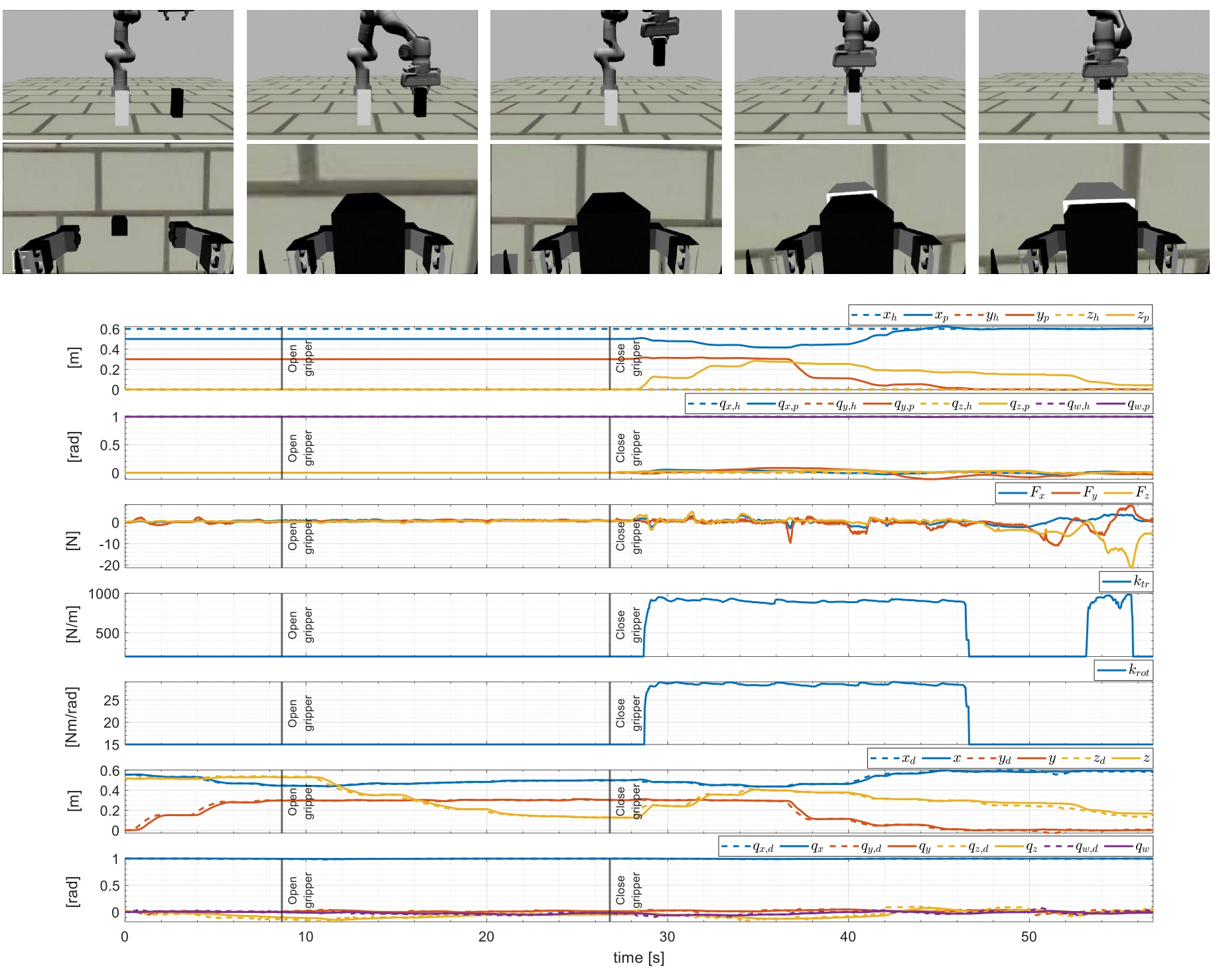}
    \caption{Top: excerpts of the visual feedback gotten by the human expert during the experiment. Bottom: plot depicting (from top to bottom) hole and peg positions, hole and peg orientations, estimated external forces at the robot end-effector, translational Cartesian stiffness, rotational Cartesian stiffness, desired and current end-effector position, and, desired and current end-effector orientation.}
    \label{fig:expert_demo_experiment}
    \vspace{-0.4cm}
\end{figure*}

\section{Experiments and Results}
\label{sec:experiments_results}

In order to validate the designed framework, we demonstrate that a human expert is able to fulfill a highly interactive task having well-defined impedance requirements when using our system. Since our framework is thought for collecting data for learning applications, we perform the experiment in simulation (gazebo\footnote{\url{https://gazebosim.org/home}})~\cite{choi2021use}.

\subsection{Experiment}

The experimental setup can be viewed in Fig. \ref{fig:data_collect_sys_design}(b)-(c). A human expert is asked to perform a peg-in-hole task fulfilling specific impedance requirements. We use simulated models of the RGB-D Realsense d435 cameras, one mounted on the wrist of the robot and one fixed in front of the robot workspace using OpenVico~\cite{fortini2022open}. Peg and hole are parallelepipeds of dimensions 50$\times$50$\times$150mm and 56$\times$56$\times$150mm (width $\times$ depth $\times$ length), respectively. The impedance requirements are as follows:
\begin{enumerate}
    \item Keep a low impedance when reaching the peg in order to reduce the impact due to possible collisions.
    \item When carrying the peg, increase the impedance in order to reject tracking errors due to the object's weight.
    \item When carrying the peg close to the hole, keep a low impedance to favor alignment and to avoid the arising of high contact forces due to contact between the peg's and the hole's sides.
    \item When inserting the peg inside the hole, increase the translational impedance to overcome friction and enable insertion while keeping low rotational impedance to avoid rotating the peg when it is inside the hole.
\end{enumerate}

\subsection{Results}
Fig. \ref{fig:expert_demo_experiment} depicts the results of the experiment. The human expert completes the task successfully, as shown by the hole and peg poses. In addition, the impedance requirements are fulfilled. Despite the misalignment between the peg and the hole at the beginning of the insertion, the interaction force does not grow too high thanks to the low impedance. Later, the high translational impedance enables the completion of the insertion by easing the sliding of the peg into the hole. Since the translational impedance is increased along all directions, the contact force between the peg and the hole grows.

\section{Conclusion}
\label{sec:discussion_conclusion}

We have shown that our framework allows us to readily provide the successful demonstration of a peg-in-hole task including impedance requirements. The successful completion of the task implies that the tracking performance of our system is sufficiently precise for teleoperating the robot, recalling that the human is able to compensate for small tracking errors if provided with visual feedback. Datasets of these demonstrations have the potential to merge the RL approaches considered in this paper with an impedance-based action space formulation like the VICES proposed in~\cite{martin2019variable}. 

We have made our system accessible to the research community hoping that it can advance research in the field of robotics for learning contact-rich manipulation tasks. The cost of our system ranges between 200\$ and 400\$ depending on the type of camera used (which is not required to provide depth information) and excluding the cost of the computer. Thus, it is much more cost effective and affordable than other commercial MoCap systems ($\gtrapprox 30000$\$) or VR devices ($\gtrapprox 3000$\$). In future works, we plan to use our framework to collect expert demonstrations to learn contact-rich tasks using the RL approaches considered in this work.

\bibliographystyle{IEEEtran.bst}
\bibliography{biblio.bib}

\begin{thebibliography}{10}
\bibitem{doi:10.1177/0278364912464668}
A.~Ajoudani, N.~Tsagarakis, and A.~Bicchi, ``Tele-impedance: Teleoperation with
  impedance regulation using a bodymachine interface (bmi) applied to the
  master operators end-effector. the concept is demonstrated in two
  experiments, namely a peg-in-the-hole and a ball-catching task, which
  illustrate complementary aspects of the method. the performance of
  tele-impedance control is assessed by comparing the results obtained with the
  slave arm under either constantly low or high stiffness.''
\end{thebibliography}

\end{document}